# The Biological Concept of *Neoteny* in Evolutionary Computation - Simple Experiments in Simple Non-Memetic Genetic Algorithms

**Vitorino Ramos**

CVRM - Instituto Superior Técnico,
Av. Rovisco Pais, Lisboa, PORTUGAL
vitorino.ramos@alfa.ist.utl.pt
http://alfa.ist.utl.pt/~cvrm

**Abstract:** *Neoteny,* also spelled *Paedomorphosis*, can be defined in biological terms as the retention by an organism of juvenile or even larval traits into later life. In some species, all morphological development is retarded; the organism is juvenilized but sexually mature. Such shifts of reproductive capability would appear to have adaptive significance to organisms that exhibit it. In terms of evolutionary theory, the process of *paedomorphosis* suggests that larval stages and developmental phases of existing organisms may give rise, under certain circumstances, to wholly new organisms. Although the present work does not pretend to model or simulate the biological details of such a concept in any way, these ideas were incorporated by a rather simple abstract computational strategy, in order to allow (if possible) for faster convergence into simple non-memetic Genetic Algorithms, i.e. without using local improvement procedures (e.g. via *Baldwin* or *Lamarckian* learning). As a case-study, the Genetic Algorithm was used for colour image segmentation purposes by using *K*-mean unsupervised clustering methods, namely for guiding the evolutionary algorithm in his search for finding the optimal or sub-optimal data partition. Average results suggest that the use of *neotonic* strategies by employing juvenile genotypes into the later generations and the use of linear-dynamic mutation rates instead of constant, can increase fitness values by 58% comparing to classical Genetic Algorithms, independently from the starting population characteristics on the search space.

**Keywords:** Genetic Algorithms, Artificial *Neoteny*, Dynamic Mutation Rates, Faster Convergence, Colour Image Segmentation, Classification, Clustering,

## 1. Incorporating *Neoteny* into Genetic Algorithms

Evolution is carried out by a process dependent on mutation and natural selection. Expositions of this thesis, however, tend to overlook the fact that mutation occurs in the genotype, whereas natural selection acts only on the phenotype, the organism produced. It follows from this that the theory of evolution requires as one of its essential parts a consideration of the developmental or epigenetic processes (due to external not genetic influences) by which the genotype becomes translated into the phenotype. Natural selection as visualised by *Darwin*, results in the production by one generation of offspring that are able to survive and reproduce themselves to form a further generation. The time unit appropriate to natural selection is therefore the generation interval. There will always be some natural selective pressure for the shortening of the generation interval, simply out of a natural economy, and for an increase of the number of offspring produced by any reproducing individual. One of the ways in which such an increase could be assured would be the lengthening of the reproductive phase in the life history; another would be an increase in the number of offspring produced [14,15,43].
There are, of course, not only natural selective pressures that operate. It is clear enough that, in evolution, they have often been overcome by other pressures. There is another natural selective of more general importance. This is the pressure to restrict the length of the reproductive period, and indeed to remove reproductive individuals, in order to make room for the maturation of a new generation in which new genetic combinations can be tried out for their fitness. A species whose individuals were immortal would exhaust its possibilities for future evolution as soon as its numbers saturated all the ecological niches suitable for its way of life. Death is a necessary condition for the trying out of new genetic combinations in later generations. It is usually brought about, in great part at least, by combinations of two processes: restriction of the period of effective reproduction to a certain portion of a life history, and as a necessary condition of this, the absence of natural selection for genetic mutations that would be effective in preserving life after reproduction has ceased. Such types of development offer the possibility of changing the relative importance of various stages in relation to the exploitation of resources and reproduction by the species. There are, for instance, many types of animals (particularly insects; e.g. termites) in which nearly the whole life history is passed in a larval stage in which most of the feeding and growth of the organism is carried out, the final adult stage being short and used almost entirely for reproduction (in simple terms, they can be seen as genetic material *carriers*). Another evolutionary strategy has been to transfer the reproductive phase from the final stage of the life history to some earlier larval stage. This again has occurred in certain insects. If such a process is carried to its logical evolutionary conclusion, the final previously adult stage of the life history may totally disappear, the larval stage of the earlier evolutionary form becoming the adult stage of the later derivative of it. It has been suggested that such process of *Neoteny* (i.e., the retention of juvenile characteristics in adulthood - the term was coined by *Kollman*) have played a decision role in certain earlier phases of evolution, evidence of which is now lost. It has been argued also, that the whole vertebrate *phylum* may have originated from modifications of one of the larval stages of an invertebrate group [43].

As his known, the process of diversity loss in genetic algorithms is often the cause of premature convergence, and as a consequence, the early convergence to an inferior local maximum. A large number of existing techniques are used to maintain diversity in *Genetic Algorithms* (GA, [11,22,29]). These include maintaining large population sizes, employing low reproductive or parent-selection pressures, applying mutation to the genotype, restarting the GA with new random genotypes, employing parallel populations (with occasional interchanging of fit chromosomes between populations), and niche-formation techniques. The present approach uses the concept of *Neoteny*. This last strategy was then incorporated in simple non-memetic genetic algorithms, by simply preserving some individuals in the earlier generations (using elitism), and by randomly re-injecting this genetic material into the later generations, allowing for substantial increases in diversity, and (as it seems) for an appropriate balance between exploration and exploitation of the search space. Some questions, however, should be discussed. For instance, at which period in the whole evolutionary process should this *neotonic* individuals be captured, how many should be thrown in (in the later generations), and when thus this *throwing* process should start? Section 7 is dedicated to those questions. Finally, and in order to study the impact of such abstract concept, yet computationally possible by using sequential incorporations of older genetic material, a difficult combinatorial problem was chosen: colour image segmentation.

## 2. Colour Image Segmentation

Image segmentation is a low-level image processing task that aims at partitioning an image into homogeneous regions [18]. How region homogeneity is defined depends on the application. A great number of segmentation methods are available in the literature to segment images according to various criteria such as for example grey level, colour, or texture. This task is hard and as we know very important, since the output of an image segmentation algorithm can be fed as input to higher-level processing tasks, such as model-based object recognition systems. Recently, researchers have investigated the application of genetic algorithms [11,22,29] into the image segmentation problem. Probably the most extensive and detailed work on GAs within image segmentation is that of *Bhanu* and *Lee* [6]. Many general pattern recognition applications of this particular paradigm can also be found in [31]. One reason (among others) for using this kind of approach is mainly related with the GA ability to deal with large, complex search spaces in situations where only minimum knowledge is available about the objective function. For example, most existing image segmentation algorithms have many parameters that need to be adjusted. The corresponding search space is in many situations, quite large and there are complex interactions among parameters, namely if we are seeking to solve colour image segmentation problems. For instance, this led *Bhanu et al*. [7] to adopt a GA to determine the parameter set that optimise the output of an existing segmentation algorithm under various conditions of image acquisition. That was the case for the optimisation of the *Phoenix* segmentation algorithm [42], by genetic algorithms, implementation described also by *Bhanu* [6]. Another situation wherein GA$^s$ may be useful tools is illustrated by the work of *Yoshimura* and *Oe* [45]. In their work, the two authors formulated the segmentation problem upon textured images as an optimisation problem, and adopt GA$^s$ for the clustering of small regions in a feature space, using also *Kohonen*'s *Self-Organising Maps* (SOM). They divided the original image into many small rectangular regions and extracted texture features from the data in each small region by using the two-dimensional autoregressive model (2D-AR), fractal dimension, mean and variance. In other example, *Bhandarkar et al.* [5] defined a multi-term cost function, which is minimised using a GA-evolved edge configuration. The idea was to solve medical image problems, namely edge-detection. In their approach to image segmentation, edge detection is cast as the problem of minimising an objective cost function over the space of all possible edge configurations and a population of edge images is evolved using specialised operators. Results comparable with those obtained using *Simulated Annealing* (SA) are reported. Fuzzy GA fitness functions were also considered by *Chun* and *Yang* [10], mapping a region-based segmentation onto the binary string representing an individual, and evolving a population of possible segmentations. Other implementations include the search of optimal descriptors to represent 3D structures [16], or the optimisation of parameters in GA hybrid systems [32] - in this last case, for finding the appropriate parameters of recurrent neural networks to segment echocardiographic images. GA applications within elastic-contour models are also possible to find. *Cagnoni et al.* [9] develop a GA based on a small set of manually-traced contours of the structure of interest (anatomical structures in 3D medical data sets). As putted by the authors, the method combines the good trade-off between simplicity and versatility offered by polynomial filters with the regularisation properties that characterise elastic-contour models. Another very interesting work, is that one of *Andrey* [1]. The image to be segmented is considered as an artificial environment wherein regions with different characteristics according to the segmentation criterion are as many ecological niches. A GA is then used to evolve a population of chromosomes that are distributed all over this environment. Each chromosome belongs to one out of a number of distinct species. The GA-driven evolution leads distinct species to spread over different niches. Consequently, the distribution of the various species at the end of the run unravels the location of the homogeneous regions on the original image. Because the segmentation progressively emerges as a by-product of a relaxation process [12] mainly driven by selection, the method has been called *Selectionist Relaxation*. In model designing terms, this last approach is indeed very close to that one presented by *Ramos* in [36], using artificial ant colonies and *Mathematical Morphology* (MM, [41]) extracted features. Approaches based on *Koza*'s *Genetic Programming* paradigm (GP, [27]), i.e., genetic algorithms used for finding appropriate algorithm structures and strategies, were also applied in image segmentation. *Poli*'s GP work [33], is perhaps one of the most interesting to follow, due to is simplicity. Finally, a fairly comprehensive review of other GA approaches in image processing is available in [8] - references include, animation, classification, feature extraction, filtering, image analysis, image processing, pattern recognition and naturally, image segmentation.

## 3. Genetic Clustering in Image Segmentation

As putted by *Andrey* [1], whether the GA is used to search in the parameter space of an existing segmentation algorithm [7], or in the space of candidate segmentations [5], an objective fitness function, assigning a score to each segmentation, has to be specified in both cases. However, evaluating a segmentation result is itself a difficult task. To date, no standard evaluation method prevails [47], and different measures may yield distinct rankings [46] (as an aside note, the present author is nowadays developing image noise measures by MM [37], allowing for instance, their use in image filtering design by GA$^s$). One possible criterion is to think of homogeneous regions as the result of any appropriate and optimised clustering process, within the image feature space. Applications of GA$^s$ in clustering and grouping problems are intensively described in [19]. In the present approach, grey level intensities of RGB image channels are considered as feature vectors, and the *k*-mean clustering model (*J.MacQueen*, 1967) is then applied as a quantitative criterion (or GA objective fitness function), for guiding the evolutionary algorithm in his appropriate search. Since the *k*-mean clustering model allows to minimise the internal feature variance of each colour cluster (or the maximisation of the variances between different colour clusters [35]), *natural* and homogeneous clusters can emerge if the GA is properly coded. In other words, the image segmentation problem is simply reformulated as an unsupervised clustering problem, and genetic algorithms are then used for finding the most appropriate and natural clusters. Since the clustering task can not be successfully applied within the image 2D space itself (e.g., similar pixels can be very far apart) the problem is coded within another space - that one of their colour features - 3D (grey level intensities, for the three channels). By this reformulation, one can in fact guarantee that similar pixels will belong to the same colour cluster. Preliminary efforts for this overall approach were designed on a previous realisation by *Ramos* in 1997 [35,38].

**4. *K*-Means Clustering Model**

*K*-Means clustering models were introduced in 1967 by *J. MacQueen*, and they are considered as an unsupervised classification technique. Once the method uses a minimum distance criteria, many authors consider this approach in many ways similar to the *k-nearest neighbour rule* method (*k*-NNR; there is also some similarities with the *Kohonen* LVQ method). In *MacQueen* terms however, *k* stands for the number of clusters searched by the model (and given as an input). All the strategy undergoes the minimisation of the expression (1). If one admits the partition of a *p*-dimensional space into *c* clusters (*c* colour classes), where *n* samples exists (*n* points characterised by *p* features), and being $C_i$ each cluster *i* centre ($i = 1,…,c$), and $X_j$ representative of each sample *j* co-ordinates ($j = 1,…,n$), where $u_{ij}$ represents the hypothetical belonging of sample *j* into cluster *i* (i.e., $u_{ij} = 1$ if *j* belongs to cluster *i*; $u_{ij} = 0$ if *j* belongs to any other cluster different from *i*). Naturally for the present case, the idea is to compute the colour cube partition minimising *J* by using genetic algorithms. We then have *J* as Eq. 1:

$$u_{ij} = 0,1$$
$$u_{ij} \in U_{c \times n}$$
$$\sum_{i=1}^{c} u_{ij} = 1$$
$$\sum_{i=1}^{c} \sum_{j=1}^{n} u_{ij} = n \quad (1)$$
$$C_i = \frac{1}{n_i} \sum_{j=1}^{n} X_j$$
$$\min J = \sum_{i=1}^{c} \sum_{j=1}^{n} u_{ij} \|X_j - C_i\|^2$$

**5. Genetic Implementation**

The above minimisation is then based on the different belonging combinations, of all points in the feature space. Naturally that, such task will be simply if the number of colours in one image to segment is low; however for high number of points in this 3D colour space (i.e., the different number of colours) this minimisation is hard to compute, since the combinatorial search space becomes very large. However, the partition of this 3D histogram into different clusters, must take in account the value of each point (that is, his frequency for a given RGB point). By this, the minimisation described in section 4, suffers a little modification (the method becomes weighted by this frequency *f*, since the number of colours of any RGB point are an important information in the overall process). The above *J* expression then becomes (Eq. 2):

$$\min J = \sum_{i=1}^{c} \sum_{j=1}^{n} u_{ij} f_j \|X_j - C_i\|^2 \quad (2)$$

Another important issue in the GA implementation is the problem's genetic coding. In order to do it appropriately, each chromosome codes the binary values of $u_{ij}$. However, to improve the GA search time and since the number of different colours in one colour image can be high, each 3D feature colour was submitted to a pre-partition. By this pre-procedure, the combinatorial search space is reduced, as also as the number of bits in each GA chromosome. That is, each 3D colour cube (with side 256 - 8 bit images) that could represent up to $56^3$ colours, was reduced to a maximum of 512 points (i.e., 512 small cubes with side 32). In other words, all RGB points that fall into a small cube are agglomerated, being the new point represented by the centre of this small cube, and his frequency being equal to the sum of all frequencies of those points.

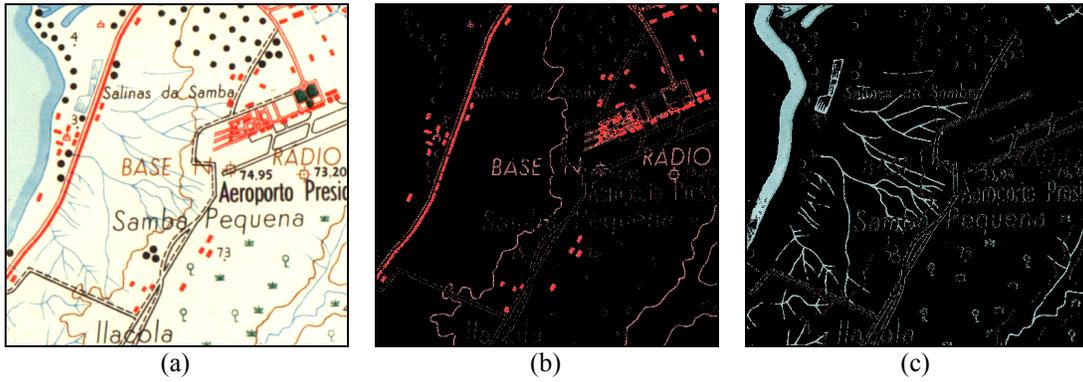

(a) (b) (c)

Figure 1 - (a) Original *Luanda* (Angola) Colour Map ([8º 50' S; 13º 15' E], from Nº89, 1:100.000, Aerial Photo – 1956 / Published – 1960) and as an example (b,c) the respective $3^{rd}/5^{th}$ colour clusters (pointed by the GA) segmenting roads, some buildings, names, topographic lines (in red-b), and rivers, lagoons, and water lines (in blue-c).

## 6. Testing and Discussing Dynamic Mutation Rates

Since the search space can be huge for similar applications (consider for example, satellite images or normal images at higher resolutions), and in order to speed-up the GA convergence (if possible), some experiments were conducted with dynamic mutation rates (i.e. time-dependent). As pointed by *Rudolph* [39] in 1994, one possible route to achieve global optimal convergence might be the introduction of time varying mutation *and* selection probabilities. *Rudolph* suggests to use two simultaneous strategies instead of one, referring the work of *Davis* (1991, [13]), where it has been shown that the introduction of time varying mutation probabilities alone does not help. Anyway, all experiments were conducted in one-point crossover genetic algorithms ($p_c$ = constant = 0.8), with 100 individuals (each pair of individuals selected via roulette wheel selection and windowing scaling, yields two new individuals), and within 3000 generations. Each individual was represented by a binary vector of length $n$ = 531 (each 3 bit can code up to 8 colour clusters, although only 6 are needed, since only 6 prominent colours are present in this maps / 177 colour small cubes present). In these conditions, each generation $g$ takes on average 0.0693 seconds (PENTIUM II - 333MHz / 128Mb RAM), which means about 3.5 minutes on 3000 generations (except for test #9, $g_{max}$ = 6000 - see table 1 / image with $500^2$ pixels and 214385 different colours - fig.1a). Then, 2 tests were run with constant mutation rates $p_m$ = 0.15 (table 1 / column D=*C*), 8 with linear-dynamic mutation rates (table 1 / column D=*LD*), and finally 25 with quadratic-dynamic mutation rates (table 1 / column D=*QD*). *C*, *LD* and *QD* tests can be expressed by the following mutation rate expressions:

- $C \Rightarrow p_m$ = 0.15; $g \in$ [0,3000]
- $LD \Rightarrow p_m$ = 0.15 (g=0) / $p_m$ = 0.15/g ; $g \in$ [1,100] / $p_m$ = 0.0015 ; $g \in$ [101,3000]
- $QD \Rightarrow p_m$ = 0.15 (g=0) / $p_m$ = $0.15/g^2$ ; $g \in$ [1,100] / $p_m$ = 0.000015 ; $g \in$ [101,3000]

Many other functions were tried, some of them inspired on *Simulated Annealing* methods (SA, [28]) or in variants of it (e.g. *Adaptive Simulated Annealing*, *Re-Annealing*, *Quenching*, [24,25,26]), as the present problem seems similar [13]. In fact, both methods are applied in search-combinatorial-optimization problems, and both start from random points in the search space. Particularly interesting in the present case is that, the mutation rate in $GA^s$ can be seen as the temperature parameter in $SA^s$ (they both affect the convergence of the respective strategy and the balance between an appropriate exploring/exploiting character of the algorithm). Similarly, scheduling temperatures in $SA^s$ (one of the most difficult problems to solve for this method) can be seen as the implementation of dynamic mutations on $GA^s$. Surprisingly (and even if several SA temperature scheduling rates were tried, generally of logarithmic or exponential nature [23,24,25,26,34,44]), the GA mutation settings that yields the best results were always the simplest ones (i.e. *LD* and *QD* - see table 2 for average results). Another fact, seems to be that the best dynamic rate should change with the starting population (compare for instance tests #2,11 and #3,12), suggesting that possibly the optimal mutation probability depends on the search landscape, the GA coding (introducing or not a multi-optimisation problem and eventually several genotype mappings to the same phenotype), and finally on the objective function itself. All the previous results appear to be in some accordance with those from *Bäck* [2,3,4] and *Mühlenbein* [30] (namely, in the hyperbolic nature of the functions used). Independently of each other, the two authors investigated in 1992, the optimal mutation rate for a simple (1+1)-algorithm (a single parent generates an offspring by means of mutation and the better of both survives for the next generation) with the objective function $f(x)=\sum_{i=1}^{n} x_I$ ("counting ones"). As putted by *Bäck* [4] the optimal mutation probability depends strongly on the objective function value $f(x)$ and follows a hyperbolic law of the form $p_m = (2.(f(x)+1)-n)^{-1}$. In order to model the hyperbolic shape of the last equation, independently of the objective function, *Bäck* used a time-dependent mutation rate $p_m(g)$ (where $n$ denotes the chromosome length, and $T$ a given maximum of generations $g$). From the condition $p_m(T-1)=1/n$, the hyperbolic formulation $p_m = (a+b.g)^{-1}$ then yields (Eq.3):

$$p_m(g) = \left( p_m^{-1}(0) + \frac{n - p_m^{-1}(0)}{T-1} \cdot g \right)^{-1} \qquad (3)$$

There is however, at least one substantial difference. As mentioned by *Bäck* based on his own research and on *Muhlenbein*'s work, practical applications of genetic algorithms often favour larger or non-constant settings of the mutation rate, and the optimal mutation rate schedule analysis for a simple objective function provides a good confirmation of the usefulness of larger, varying mutation rates (in classical approaches they are generally $p_m \in [0.001, 0.01]$, see [11,22]). For these reasons, *Bäck* imposed $p_m(0)=½$. However, comparing the GA efficiency based in *Bäck*'s function (Eq. 3 / with $p_m(0)= ½$ or $p_m(0)=0.15$ / $T=3000$ / $n=531$), with the *LD/QD* functions, we come up with significant differences (tests #2,3,4,5). These results (although, they are statistically insufficient) probably point that optimal dynamic mutation rates should also be characterised in function of the problem's search landscape (which are manifestly different - "counting ones" *versus* Eq. (1)).

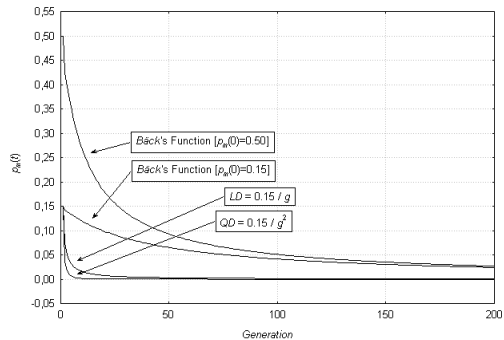

Fig. 2 - Comparing *Bäck*'s [2,4] dynamical mutation rate function (with $p_m(0)= ½$ and $p_m(0)=0.15$) and *LD, QD* functions, for $g \in [0,200]$.

*Bäck* followed this route [4], adapting the mutation rates according to the topology of the objective function, using the principle of strategy parameter self-adaptation as developed by *Schwefel* [40,3] for *Evolution Strategies* (ES), or similarly and independently by *Fogel* [20] for *Evolutionary Programming* (EP, [21]). These models, however, were not applied or have been analysed in the present framework; instead, a novel approach was considered: artificial *neoteny* (*aNeoteny*).

## 7. Implementing and Testing Artificial *Neoteny* (*aNeoteny*)

In order to implement *aNeoteny*, the preservation of older genotypes is a key-aspect. In general, this preservation was possible through capturing elitist individuals (*neotonic* individuals) from generations $g=0$ till $g=100$ (one per generation), and *throwing* them randomly into later generations (i.e a randomly individual give his place in the population array, to one randomly chosen *neotonic* individual, generally for $g \in [1000,3000]$). Some questions, however, seems to be pertinent. For instance, at which period in the whole evolutionary process should this *neotonic* individuals be captured, how many should be thrown in (in the later generations), and when thus this *throwing* process should start? In order to answer this questions and to evaluate the possible contribution of *Neoteny* in the GA fitness convergence, several tests (38) were conducted (table 1). These tests can be roughly classified into six groups. The first group include tests #1 to #8, and his purpose was to evaluate and compare the GA perfomance for different types of mutation with or without the implementation of neoteny (#7,8) for the same random seed. The second group (tests #10-14) aims to evaluate the same effect but now for a different starting population (the nature of different starting populations can be analysed, in terms of fitness, by table 3). Since the results of the first group suggest the use of neotonic strategies, while the second achieves roughly the same fitness values by incorporating only dynamic mutation rates, the third group (tests #15-25) was dedicated to evaluate if neotonic injection of genotypes could achieve the same results when incorporating that material at different generation intervals (i.e., at different evolutionary periods). Following the same concern, tests #26-30 (fourth group) analyses the effect on the average number of thrown neotonic individuals. The fifth group (tests #31-34) concerned the generation interval where neotonic individuals should be captured, and finally the sixth group (tests #35-38) analyses the effect of re-injecting one neotonic individual simultaneously with one complete random created individual. Average results, for different starting populations and strategies, can be found at table 2, and table 3 presents the random seed effect on some characteristics for these different starting populations used. Finally, fig. 3 presents the convergence of some GA strategies for each generation.

## 8. Conclusions and Future Work

The above strategy was applied in colour maps (214385 different colours - see figure 1). For different image applications (colour skin mark segmentation / ornamental stone segmentation) see (*Ramos* and *Muge*, [38]). Since the colour maps have 6 prominent colours, the aim was to search for 6 colour clusters. Overall results point to highly satisfactory results namely for the segmentation of ornamental stones (29349 different colours) and for the case of human skin mark segmentation (303 colours). There is however some problems with the colour map examples. The main reason is that the problem is by itself difficult (with large combinatorial search spaces), and that the pre-partition tends to reduce the discriminatory power of the overall strategy. This is mainly observed within pixels that form bounds of any important colour object. Image acquisition with low resolutions interpolates somehow their grey level intensities into intermediate values (between inner and outer bounds), and the result (with pre-partition) is significantly altered, since similar pixels can belong to different small cubes (naturally with low probability). However if the number of this kind of pixels is high, the strategy tends to create himself another cluster. On the other hand, and by observing the GA performance (*J* value) in each generation, we can conclude that similar results can be achieved with half of the generations run (3000), since after this point, *J* values are increasing very slow (compare tests #2,9 - a double value of generations adds around 0.45% in the fitness value, which is counter-productive). Future work includes three main lines. First, to study the cluster relations (clouds of points) for each segmentation problem. This can bring useful information into the GA approach, and simply geodesic neighbourhood relations can be computed by using *Mathematical Morphology* [41] on the 3D-colour cube. Second, more relevant evaluating methods for image segmentation must be studied. In this topic, *Zhang*'s work [47] should be followed if possible. The present authors are making nowadays, preliminary attempts, even if they are related with image noise [37]. Significant improvements on the automatic design could also be achieved by using ISODATA models - since the number of clusters can be automatically chosen by the hybrid search model. Finally, the present approach can be used for general classification purposes, since the *K*-Means clustering method (the present GA objective function) can be formulated for several dimensions and different types of features.

Regarding the neotonic strategies, and by analysing the results of tests #1 to #14 (first and second test groups), it is clear that the strategy of implementing neotonic strategies and dynamic mutation rates can yield substantially (around 58%) the fitness values for the same number of generations (3000), comparing to the use of constant mutation rates (see also table 2). The best result was achieved by using dynamic mutation rates and neotonic strategies (#6), although when we change the starting population the same result was only achieved by using non-neotonic implementations (test #12). It appears that starting populations with above-average individuals on it (see table 3 - random seeds $R=917$, $R=27$ and $R=7445$, which is the case of test group II) do not need for higher exploring natures in the search space to achieve above-average fitness, either by incorporating a slowest decay in the mutation rate (e.g. *LD* versus *QD*) or by yielding the population diversity into the later generations via neotonic strategies. In fact, they appear to achieve good results simple by exploiting the above-average fitness and schema of their population. This is probably why, at constant mutation rates, the starting population with *R*=9 (test #1) with greater average fitness, achieves better results than test #10 (*R*=7445).

| Test # | A | B | C | D | E | F | G | H |
|---|---|---|---|---|---|---|---|---|
| 1 | 9 | 3000 | 0.8 | C | 0 | - | - | 201.611623 |
| **2** | **9** | **3000** | **0.8** | **LD** | **0** | **-** | **-** | **325.528410** |
| 3 | 9 | 3000 | 0.8 | QD | 0 | - | - | 312.694066 |
| 4 | 9 | 3000 | 0.8 | B [0.15] | 0 | - | - | 203.964332 |
| 5 | 9 | 3000 | 0.8 | B [0.50] | 0 | - | - | 180.236736 |
| **6** | **9** | **3000** | **0.8** | **LD** | **1** | **[1,100]** | **[1000,3000]** | **326.426236** |
| 7 | 9 | 3000 | 0.8 | QD | 1 | [1,100] | [1000,3000] | 314.125107 |
| 8 | 9 | 3000 | 0.8 | B [0.15] | 1 | [1,100] | [1000,3000] | 207.823020 |
| **9** | **9** | **6000** | **0.8** | **LD** | **0** | **-** | **-** | **326.993288** |
| 10 | 7445 | 3000 | 0.8 | C | 0 | - | - | 191.146788 |
| 11 | 7445 | 3000 | 0.8 | LD | 0 | - | - | 306.475341 |
| **12** | **7445** | **3000** | **0.8** | **QD** | **0** | **-** | **-** | **326.549272** |
| 13 | 7445 | 3000 | 0.8 | LD | 1 | [1,100] | [1000,3000] | 308.919431 |
| 14 | 7445 | 3000 | 0.8 | QD | 1 | [1,100] | [1000,3000] | 321.010773 |
| 15 | 7445 | 3000 | 0.8 | QD | 1 | [1,100] | [500,3000] | 319.063587 |
| 16 | 7445 | 3000 | 0.8 | QD | 1 | [1,100] | [350,3000] | 320.481312 |
| 17 | 7445 | 3000 | 0.8 | QD | 1 | [1,100] | [320,3000] | 316.784335 |
| 18 | 7445 | 3000 | 0.8 | QD | 1 | [1,100] | [300,3000] | 322.772565 |
| 19 | 7445 | 3000 | 0.8 | QD | 1 | [1,100] | [285,3000] | 316.366818 |
| **20** | **7445** | **3000** | **0.8** | **QD** | **1** | **[1,100]** | **[280,3000]** | **324.908299** |
| 21 | 7445 | 3000 | 0.8 | QD | 1 | [1,100] | [279,3000] | 317.947635 |
| 22 | 7445 | 3000 | 0.8 | QD | 1 | [1,100] | [277,3000] | 318.843974 |
| 23 | 7445 | 3000 | 0.8 | QD | 1 | [1,100] | [275,3000] | 319.100290 |
| 24 | 7445 | 3000 | 0.8 | QD | 1 | [1,100] | [200,3000] | 316.244083 |
| 25 | 7445 | 3000 | 0.8 | QD | 1 | [1,100] | [150,3000] | 316.556148 |
| 26 | 9 | 3000 | 0.8 | LD | 2 | [1,100] | [1000,3000] | 319.990759 |
| 27 | 7445 | 3000 | 0.8 | QD | 1.5 | [1,100] | [280,3000] | 312.452034 |
| 28 | 7445 | 3000 | 0.8 | QD | 2 | [1,100] | [280,3000] | 311.933670 |
| 29 | 7445 | 3000 | 0.8 | QD | 3 | [1,100] | [280,3000] | 303.136676 |
| 30 | 7445 | 3000 | 0.8 | QD | 5 | [1,100] | [280,3000] | 297.281200 |
| 31 | 7445 | 3000 | 0.8 | QD | 1 | [100,200] | [1000,3000] | 317.683124 |
| 32 | 7445 | 3000 | 0.8 | QD | 1 | [100,200] | [280,3000] | 309.894177 |
| 33 | 7445 | 3000 | 0.8 | QD | 1 | [1,50] | [280,3000] | 322.543241 |
| 34 | 7445 | 3000 | 0.8 | QD | 1 | [1,30] | [280,3000] | 317.450920 |
| 35 | 9 | 3000 | 0.8 | LD | 2* | [1,100] | [1000,3000] | 323.254605 |
| 36 | 9 | 3000 | 0.8 | QD | 2* | [1,100] | [1000,3000] | 315.927842 |
| 37 | 7445 | 3000 | 0.8 | LD | 2* | [1,100] | [1000,3000] | 290.810651 |
| **38** | **7445** | **3000** | **0.8** | **QD** | **2*** | **[1,100]** | **[1000,3000]** | **326.281866** |

Table 1 - Results for 38 GA runs in terms of fitness (column H: $10^9/J$). Column A: Random seed; Column B: maximum number of generations; Column C: Crossover probability; Column D: Type of Mutation (*C*=constant=0.15, *LD* or *QD* / B = *Bäck*'s function with $p_m(0)$= ½ or $p_m(0)$=0.15); Column E: average number of *Neotonic* individuals re-injected in the generation interval at column G (*one individual completely random created re-injected with one *Neotonic* individual); Column F: Generation interval where *Neotonic* individuals were captured (one for each generation).

| GA Strategy (see Test # - Table 1) | | R=9 | R=7445 | R=917 | R=14 | R=27 | Average |
|---|---|---|---|---|---|---|---|
| C | (1,7) | 201.61 | 191.15 | 183.36 | 205.07 | 201.52 | 196.54 |
| LD | (2,8) | 325.53 | 306.48 | 275.41 | 322.07 | **314.29** | 308.76 |
| QD | (3,9) | 312.69 | **326.55** | 286.61 | 290.89 | 270.74 | 297.50 |
| LD/N | (4,10) | **326.43** | 308.92 | 285.14 | **323.07** | 313.87 | **311.49** |
| QD/N | (5,11) | 314.13 | 321.01 | **310.42** | 281.46 | 279.56 | 301.32 |
| LD/N+R | (35,37) | 323.25 | 290.81 | 284.94 | 317.57 | 312.83 | 305.88 |
| QD/N+R | (36,38) | 315.93 | 326.28 | 292.87 | 297.81 | 305.70 | 307.72 |
| Average | | 302.80 | 295.89 | 274.11 | 291.13 | 285.50 | 289.89 |

Table 2 - Analysis of different GA strategies with different starting populations (see Table 1 for similar test types / values for 3000 generations / best values for each random seed are in bold).

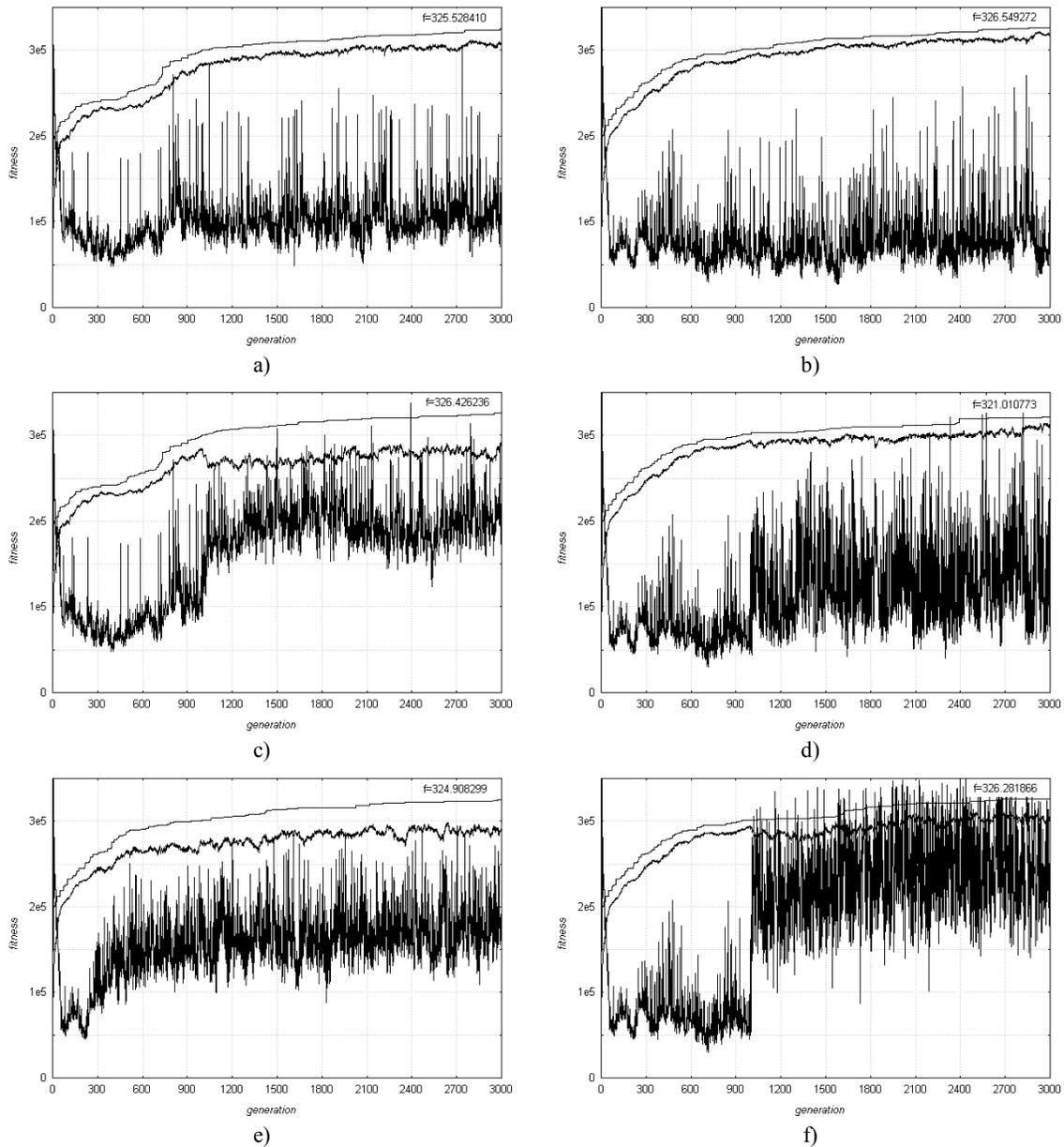

Fig. 3 - Best individual fitness ($10^9$/$J$), average fitness ($10^9$/$J$) and fitness standard deviation (x10) for each generation (100 individuals). a) *LD*, *R*=9 (test #2 - see Table 1). b) *QD*, *R*=7445 (test #12). c) *LD/N*, *R*=9 (test #6). d) *QD/N*, *R*=7445 (test #14). e) *QD/N*, *R*=7445, and incorporating *Neotonic* individuals for *g*>280 (test #20). f) *QD/N+R*, *R*=7445 (test #38).

| Random Seed | Best Fitness | Worst Fitness | Average Fitness | Std.Dev. | Sum |
|---|---|---|---|---|---|
| R=9 | 126.1 | 81.0 | 104.8 | 12.2 | 10484.0 |
| R=14 | 125.6 | 78.4 | 100.7 | 11.8 | 10068.7 |
| R=27 | 128.5 | 78.2 | 100.1 | 11.7 | 10005.4 |
| R=917 | 132.9 | 79.0 | 101.3 | 12.3 | 10125.8 |
| R=7445 | 128.5 | 79.3 | 101.9 | 12.1 | 10192.1 |

Table 3 – Random seed effect on the initial population, in terms of fitness ($10^9$/$J$) for *g*=0 (100 chromosomes)

It appears also (see tests #15-25) that under these circumstances, no optimal neotonic strategy can be found. In fact, throwing neotonic individuals at different temporal periods point that results can be different and only near fitness values could be found (test #20). However, introducing diversity by neotonic implementations and simultaneously incorporating diversity into this diversity, by adding complete random created individuals (tests #35-38) could yield the fitness values to the same level, for *R*=7445. Apparently this last argument is in contradiction with the one of the last paragraph. However, is the author belief that for some starting populations (e.g., *R*=7445) the increase of diversity (increasing the exploring capabilities of the algorithm) by neotonic strategies cannot fulfil the exploiting power of simple genetic algorithms, unless, this diversity is himself increased. In other words, for a finite number of generations and for the precedent contexts, the best convergence could only be achieved either by increasing the exploring character of the algorithm, or by increasing his exploiting character, that is, renouncing for the suppose-to-be appropriate exploring/exploiting balance. This last point suggests that probably, a diversity *critical-mass* is needed within the evolutionary process, for some starting points in the search landscape.

On the other hand, tests #26-30 suggest that no better results could be found by re-injecting more than one neotonic individual per generation. In fact, results decay has the number of neotonic individuals increases. Finally, results also change if neotonic individuals are captured in different time-windows (tests #31-34 / column F - table 1). Why the interval [1,100] for capturing neotonic individuals, and the interval [1000,3000] for throwing them appear to be optimal, however, is hard to answer. Nevertheless, it appears to be important to give to the evolutionary search some time before re-injecting neotonic individuals, i.e. some evolutionary period where genetic exploitation should be processed in the classical way. Further tests should be implemented in order to analyse this point. Finally, a note about the neotonic strategy effect on the genetic image segmentation processing. In the case of colour images, the differences between both techniques (classical *versus* neotonic) clearly affects the visual quality, namely at enhancing objects extracted (also) by the classical way.

## Acknowledgements

The author wishes to thank to *David Fogel* (Natural Selection, Inc. / USA), *Thomas Bäck* (Center for Applied Systems Analysis - CASA / Germany) and *Rajeev Ayyagari* (Indian Statistical Institute / India), for their useful references and comments on Dynamic Mutation and *Neoteny* at *COMP.AI.GENETIC* (Jan./Feb. 2000), and also to FCT-PRAXIS XXI (BD20001-99), for his PhD Research Fellowship.